\documentclass[runningheads]{llncs}

\usepackage{eccv}

\usepackage{eccvabbrv}
\usepackage{graphicx}
\usepackage{booktabs}

\usepackage[accsupp]{axessibility}  

\usepackage[pagebackref,breaklinks,colorlinks,citecolor=eccvblue]{hyperref}

\usepackage{orcidlink}

\begin{document}

\title{VQA-Levels: A Hierarchical Approach for Classifying Questions in VQA} 


\author{Madhuri Latha Madaka\inst{1}\orcidlink{0000-1111-2222-3333} \and
Chakravarthy Bhagvati\inst{1}\orcidlink{1111-2222-3333-4444}} 
\institute{University of Hyderabad, Hyderabad, Telangana, India 
\email{19mcpc02@uohyd.ac.in,chakravarthybhagvati@uohyd.ac.in }}

\maketitle

\begin{abstract}
Designing datasets for Visual Question Answering (VQA) is a difficult
and complex task that requires NLP for parsing and computer vision for
analysing the relevant aspects of the image for answering the question
asked. Several benchmark datasets have been developed by researchers
but there are many issues with using them for methodical performance
tests. This paper proposes a new benchmark dataset -- a pilot version
called VQA-Levels is ready now -- for testing VQA systems
systematically and assisting researchers in advancing the field. The
questions are classified into seven levels ranging from direct answers
based on low-level image features (without needing even a classifier)
to those requiring high-level abstraction of the entire image
content. The questions in the dataset exhibit one or many of ten
properties. Each is categorised into a specific level from 1 to
7. Levels 1 - 3 are directly on the visual content while the remaining
levels require extra knowledge about the objects in the image. Each
question generally has a unique one or two-word answer. The questions
are 'natural' in the sense that a human is likely to ask such a
question when seeing the images. An example question at Level 1 is,
``What is the shape of the red colored region in the image?" while at
Level 7, it is, ``Why is the man cutting the paper?". Initial testing of the proposed dataset on some of
the existing VQA systems reveals that their success is high on Level 1
(low level features) and Level 2 (object classification) questions,
least on Level 3 (scene text) followed by Level 6 (extrapolation) and
Level 7 (whole scene analysis) questions. The work in this paper will
go a long way to systematically analyze VQA systems.

\end{abstract}

\section{Introduction}
\label{sec:intro}

Visual question answering (VQA) is a challenging multimodal task
involving vision and language. It requires processing and
understanding the visual and semantic characteristics of an image and
the linguistic properties of a question. The correct answer then
requires combining these two with perhaps additional analysis that
lies in between these two domains. VQA is an active research area with several
systems\cite{jiang2018,kim2021,wang2022,wang2022git,li2022} developed in
the last few years often as extensions of the image captioning
systems. 

An important component of VQA research is the generation of
appropriate datasets for training as well as testing purposes. Several
VQA datasets like VQA 1.0\cite{antol2015}, VQA 2.0\cite{goyal2017},
Visual 7W\cite{zhu2016}, Visual Madlibs\cite{yu2015},
CLEVR\cite{johnson2017}, TDIUC\cite{kafle2017} are used by
researchers. Another popular dataset is MSCOCO\cite{lin2014microsoft}
which is primarily developed for image captioning and for image
descriptions.

There are many challenges associated with building a VQA dataset. The
questions have to be so framed that they have unambiguous answers. The
existing datasets provide many different answers and a VQA system is
evaluated on its answers being one of the given answers. The questions
sometimes are deliberately designed to confuse VQA
systems as in VQA\,1.0\cite{antol2015}. The questions are not
graded according to difficulty levels but classified based on common
question forms such as, ``what, how, which, when,'' \etc. Many
questions are \emph{contrived} in the sense that they are not commonly
asked by humans who are shown the image. There are also irrelevant
questions which are completely unrelated to the objects in the
image. Finally, it is not an easy task to generate questions although
attempts have been made to do so \cite{gao2015you}. The
resulting questions are usually based on low-level features such as
color, shape and positions in the image.

Humans analyze images at varying levels of complexity and
abstractions. As toddlers, color attracts our eyes, sometimes size
and shape of the objects, and as we grow older, familiarity, emotional
and higher level abstractions based upon the entire scene and not on
individual objects form the basis for our questions. Questions also
depend on the \emph{purpose} or the \emph{application} for which we
look at the image.

It is easy to recognise that humans ask a number of questions with
varying levels of complexity and familiarity when seeing an image. For
example, if shown an image of a Premier League match, a person not
familiar with the league may ask, ``What is being played?''; another
who knows football may ask, ``Which teams are playing the match?''; a
third, who is a Premier League fan may even look for a popular player
\emph{X} and ask, ``Where is \emph{X}?''

A suitable benchmark dataset must address the above issues and also be
generic without specific references to local or regional aspects
unless these are popular worldwide. In this paper, we propose a new
dataset called the \emph{VQA-Levels Dataset} that addresses many of
the issues above. All the questions in the dataset have unambiguous
answers that are mostly a single word or may be two to three word
phrases. They are organised into 7 levels of complexity with Level 1
containing the most direct questions that do not need even a
classifier to answer them correctly. Level 7 contains abstract
questions based on the entire image such as, ``Why are tomatoes being
thrown at each other?'' Each higher level, in general, requires
greater analysis than its lower levels. A pilot version with about 210
images and 751 questions is ready but the goal is to create a much
larger dataset with around 2000 questions on about 400 images. As the
size of the pilot dataset is small, it is currently usable only for
testing VQA systems rather than training.

The main contributions of our work are (a) creation of a VQA-Levels
dataset with systematic categorisation of questions into graded levels
of difficulty; (b) analysis of the results of popular VQA systems with
regards the complexity level of questions; and, (c) finding specific
areas for improvement in performance of VQA systems.

The paper is organised into 7 sections. The second section presents
some currently existing VQA datasets with question categories. Section
3 provides the intuitions and the background of VQA-Levels dataset
creation while Section 4 describes the 7 level hierarchy. Section 5
describes the VQA-Levels dataset, the experiments and results of testing
several popular VQA systems with the VQA-Levels dataset. Section 6 is
a discussion on the findings from the experiments with conclusion in
Section 7.


\section{Related Work}
\label{sec:relatedWork}
There are several datasets popular with researchers in VQA. Some of
them classify the questions into groups based on question
\emph{keywords}. 
\begin{description}
\item[COCO-QA]\cite{ren2015} have questions in four categories on
  object identification, number, color and location. All the questions
  have one word answers. It has 123287 images and 117684 questions.
\item[VQA 1.0, VQA 2.0]\cite{antol2015,goyal2017} Both these
  annotated datasets group questions based on the starting words of
  the question like ``What is'', ``Is there'', ``How many'',
  ``Which'', ``What time'', ``What sport is'' , \etc. VQA 2.0 is one
  of the most comprehensive datasets for VQA with many categories of
  images and questions.
\item[CLEVR]\cite{johnson2017} asks questions based on identification
  of attributes,  comparison of attributes, number of objects,
  existence and integer comparison of objects. It has 100,000 images
  and more than 850,000 questions. 
\item[Visual7W]\cite{zhu2016} has seven kinds of queries starting with
  ``what, how, where, who, when, why''  and ``which.''
\item[Visual Genome]\cite{krishna2017} has six question types ``what,
  why, where, who, when'' and ``how'' in a multiple
  choice setting.
\item[TDIUC]\cite{kafle2017} has eight categories of questions
  like presence of object, recognition of subordinate object, count,
  color, other attributes, recognition of activity, sport recognition
  and positional recognition. It has about 1.6 million questions on
  170,000 images.
\end{description}
The above datasets are all extremely large in size but the questions
are not organised systematically into categories with increasing
complexity. The question categories are non-hierarchical and are also
based almost completely \emph{only} on the visual content in the
image. A human being, on the other hand, asks and answers questions
based on the \emph{semantics} and \emph{object properties}. For
example, in a mixed double match showing three players -- two males
and a female -- a question on the gender of the fourth unseen player
in the image is easily answered by a human but such questions are not
present and categorised in the above datasets. Questions that require
external knowledge are also not systematically identified in these
datasets. An example is, ``Which two fruits in the image are similar
in taste?'' Taste has no visual counterpart.

The VQA-Levels dataset systematically categorises the questions into 7
levels (see \cref{7leves}). These questions are also based on
external knowledge which a human is generally expected to have. An
example is, ``How many real towels are there in the image?'' with the
image having a mirror reflection of a single towel.

\section{VQA-Levels Dataset}
\label{sec:VQALevels}

The intuition behind the creation of VQA-Levels dataset is
three-fold. The first, to create a dataset containing questions at
different levels of complexity. The second, to have questions that a
human being is likely to ask and which may go beyond \emph{only} the
visual content in the image. The third, to have questions that can be
answered in an unambiguous manner in either one word or two to three
word phrases.

The first task is, therefore, to identify levels of complexity in a
systematic manner. The inspiration came from Marr's Theory of three
levels of vision\cite{marr2010} and content-based image retrieval
systems (CBIR). Marr's theory states that \emph{low-level} features are
first extracted from images and are used to form an intermediate
$2\frac{1}{2}D$ sketch that divides image into segments with shape and
3D position information. The third or high-level assigns semantic
labels to the different regions in the image and recognises them as 3D
objects. CBIR systems classified questions into categories based on
visual content, object properties and similarities and semantic
contents.

Questions in the VQA-Levels dataset are human oriented. Our aim is to
ask questions which humans would ask when they look at an image rather
than to focus on VQA systems, their architectures and
algorithms. There is no attempt to ask questions that are designed to
confuse the system as in VQA 1.0\cite{antol2015}. All questions in
VQA-Levels are precise and have mostly only one correct response,
which will aid in evaluating the performance more effectively. This
also led us to decide that automatic generation of questions based on
image descriptions will not be used in the initial phase.

The questions are framed so that they can be answered
unambiguously. An answer was given by the person who gave a
question. Multiple students then answered the same question
independently and were given the option of either sticking with their
answer or revert to the answer given by the questioner. In this
process, questions which had multiple answers were excluded from the
dataset. 

These three main features resulted in a dataset containing 7 levels of
complexity. The first two levels L1 and L2 involve low-level features
such as color, shape, texture, \etc. and object
identification. Levels L4 to L7 deal with symbolic and semantic
information, similarities and dissimilarities in object properties and
such combinations of visual content and semantics. Level L3 will be
described later in this section.

\subsection{Visual and Semantic Properties}
The properties used in creating the VQA-Levels dataset are:
\begin{description}\label{objprops}
\item[Intrinsic Properties:] An object has many properties, e.g a
  bicycle has two wheels, a bottle can contain solid or liquid
  objects, etc. A question such as ``Which object in the image can be
  used to hold water?'' is an example question based on certain
  intrinsic properties of an object labelled by a classifier.
\item[Emotions:]    Emotions can be specific to a single object in the
  image or at the entire image level. For example, ``How is the person
  in red shirt feeling?''
\item[Directions and Signs:]    There may be arrows, traffic lights
  and other such objects with specific functions and a question can be
  on such functions. For example the question, ``Is it safe for the
  pedestrian to cross the road?'' in an image showing a pedestrian
  walk signal.
\item[Similarity/Dissimilarity:] Questions based on comparisons such
  as ``Which object in the image is similar in shape to a banana?''
  (banana may or may not be present in the image) or
  ``Which two fruits in the image have different tastes?'' relate to
  similarity and dissimilarity. They can be with regard to a specific
  object or can be on the entire scene.
\item[Seasons and Time of the Day:] Images often contain information
  about time and seasons. For example, a tree with colorful leaves
  indicates Fall season and a snowy scene is winter time. Blue sky
  indicates day time and long shadows may indicate either dawn or
  dusk.
\item[Distinguishable Property:] Certain objects are uniquely
  identifiable and associated with specific contexts. E.g., ``In which
  city is this photograph taken?'' for an image with Eiffel Tower in it.
\item[Functionality:] Objects have a specific purpose or role it
  serves within a system or context like clock used to show time, bell
  of bicycle to alert. For example, in an image with a man wearing a
  raincoat on a rainy day, the question ``Which object protects the
  man from rain?'' is related to functionality.
\item[Sequencing:] Order of things and events - what happened before
  and what will happen after. For example, ``Which colored ring will
  come next?'' while arranging a group of rings in descending order of
  size, ``Will the batsman hit the ball?'' from a sports image.
\item[Extrapolation and Occlusion:] A portion of an object or a scene
  is visible in the image and the question relates to the occluded or
  unseen portions. For example, ``How many wheels does the blue object
  have?'' Once the object is identified as a car, the answer is
  ``four'' even if not all wheels are visible in the image.
\item[Intangibles and Abstractions:] Certain questions may require
  analysis and knowledge of all the objects in the scene shown in the
  image. For example, ``What object are the objects in this image a
  part of?'' on an image showing the case, cap, refill and other
  parts of a pen. ``Which event is being celebrated?'', is another
  example.
\end{description}

Level L3 came out of the process of creating the VQA-Levels
dataset. The dataset was created through the efforts of about 50
students (see Section \ref{sec:expandres}). From the questions, it
became clear that any text, if present in an image, is a significant
source of information. Many questions are related to processing the text
in the images. Thus, L3 is concerned with questions on \emph{scene
  text} which may be either text, \eg signs, name-boards, \etc or
clocks, either analogue or digital, calendars, \etc. Examples of L3
questions are shown in Figures \ref{fig:L3example1} and
\ref{fig:L3example2}.

\section{Question Levels}
\label{7leves}
We have classified the questions in a VQA system into 7 levels by
combining CBIR systems approaches and Marr's theory as our
foundation. The first two levels are purely visual and may be answered
by constructing a suitable classifier (the third
level in Marr's theory). The third level in our VQA questions hierarchy
requires reading and processing \emph{scene text}. Fourth and higher
levels ask questions that require a symbolic or semantic
representation based on the object labels given by the classifier;
having only visual content is insufficient. For example, a question,
``What is the object held by the woman in the image used for?''  or
``What event is being celebrated?'' is at higher levels.
Questions at levels 4 -- 7 are based on the features described in the
previous section.

\begin{description}
\item[Level 1 (L1):] These are questions which can be answered without
  classification of objects. They query on low level features of the
  image like color and shape which do not need labeling the object to
  answer them. Examples are given in \cref{fig:L1example}.
  
\item[Level 2 (L2):] These questions can be answered on applying a
  classifier.  Deep networks trained for classification on datasets
  like ImageNet\cite{deng2009} must be able to answer these
  questions. Questions on
    \begin{itemize}
       \item object identification
       \item logo identification
       \item color, position, shape of objects (not regions)
       \item number of objects: multiple objects identified by a
          classifier that belong to either the same or different
          categories
    \end{itemize}
  come under Level 2. Examples are given in \cref{fig:L2example}.

\begin{figure}[tb]
  \centering
  \begin{subfigure}{0.22\linewidth}
    \centering
    \includegraphics[width=0.95\textwidth,height=3.4cm]{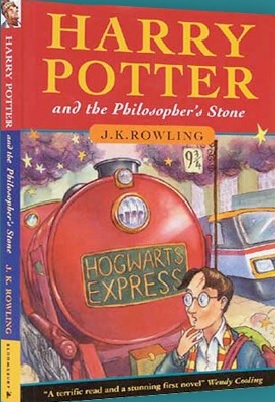}
    \vspace{\baselineskip} 
    \begin{small}
        \begin{tabular}{p{.5cm}p{2.2cm}}
            
            \textbf{Q}: & What is the dominant color in the picture?\\
            \textbf{A}: &  red\\

        \end{tabular}
    \end{small}
    \caption{}
    \label{fig:L1example}
  \end{subfigure}
  \hfill
  \begin{subfigure}{0.22\linewidth}
    \centering
    \includegraphics[width=0.95\textwidth,height=2.5cm]{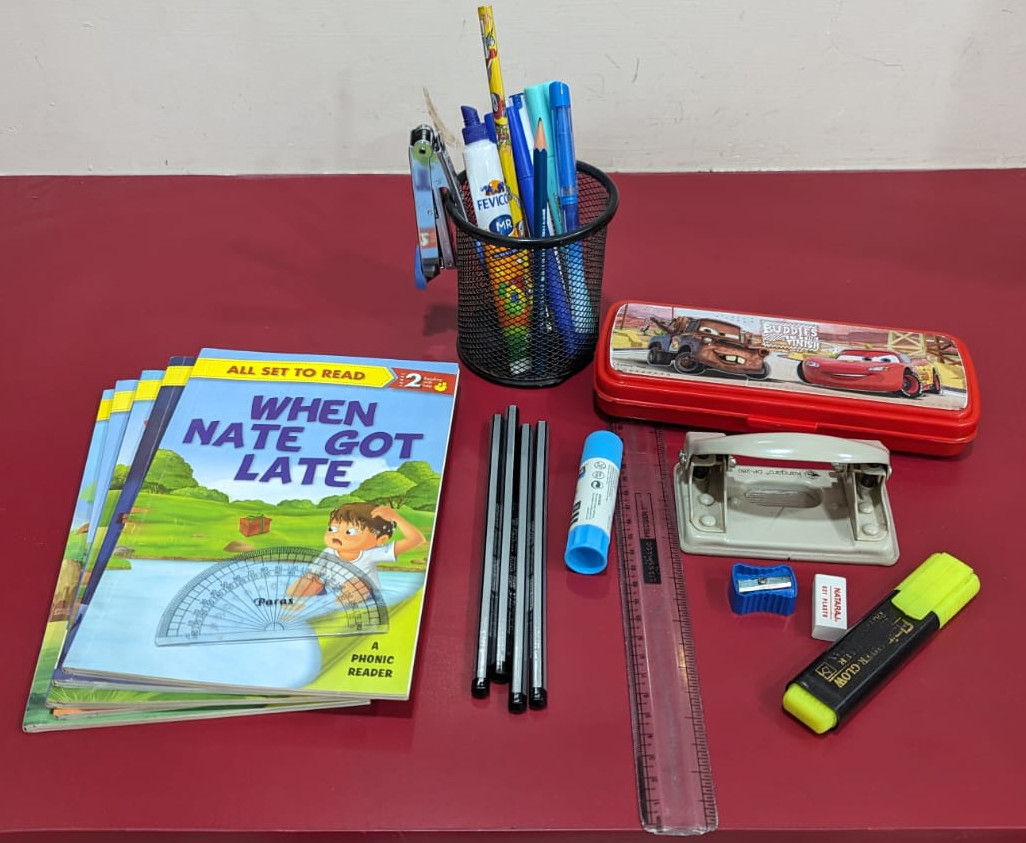}
    \vspace{\baselineskip} 
    \begin{small}
        \begin{tabular}{p{.5cm}p{2.2cm}}
           
            \textbf{Q}: & How many pencils are beside the books?\\
            \textbf{A}: &  4\\

        \end{tabular}
    \end{small}
    \caption{}    
    \label{fig:L2example}
  \end{subfigure}
  \hfill
  \begin{subfigure}{0.22\linewidth}
    \centering
    \includegraphics[width=0.95\textwidth,height=2.8cm]{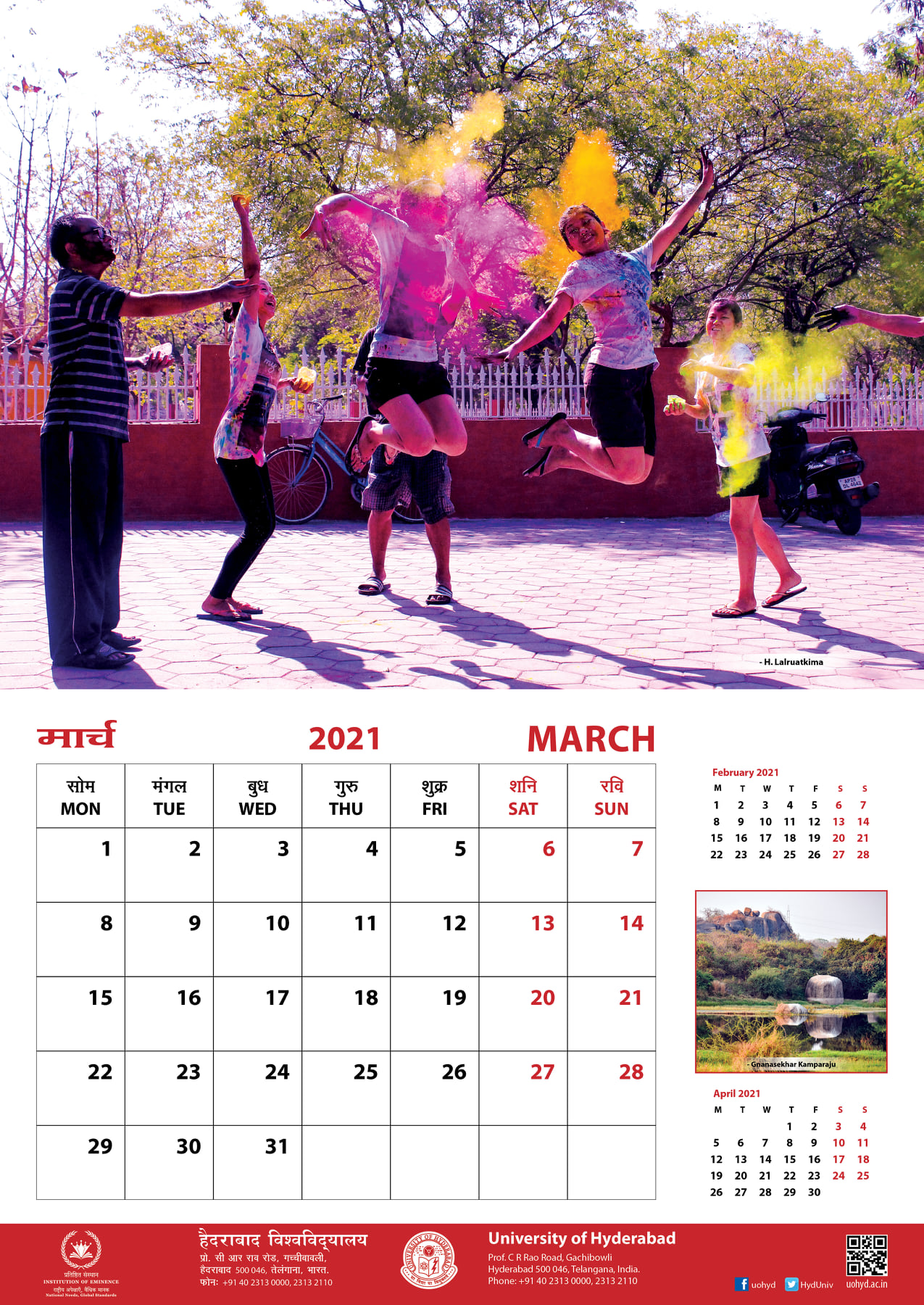}
    \vspace{\baselineskip} 
    \begin{small}
        \begin{tabular}{p{.5cm}p{2.2cm}}
            
            \textbf{Q}: & Which month and year's calendar is shown?\\
            \textbf{A}: & March 2021    \\
                       
        \end{tabular}
    \end{small}
    \caption{}
    \label{fig:L3example1}
  \end{subfigure}
  \hfill
  \begin{subfigure}{0.22\linewidth}
    \centering
    \includegraphics[width=0.95\textwidth,height=2.8cm]{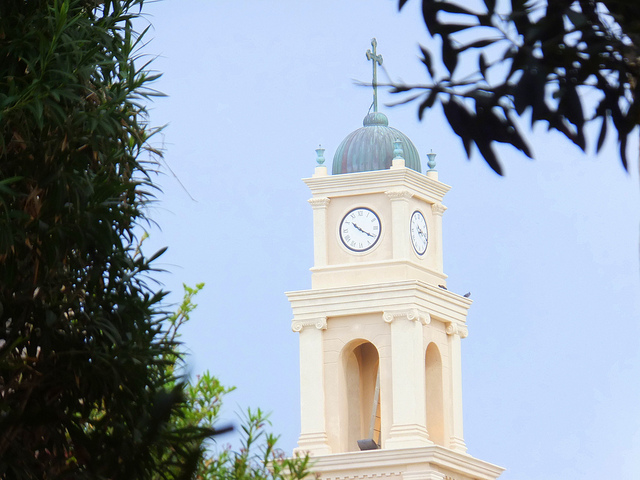}
    \vspace{\baselineskip} 
    \begin{small}
        \begin{tabular}{p{.5cm}p{2.2cm}}
            \textbf{Q}: & What time is it?\\
            \textbf{A}: & 10:20\\
            
        \end{tabular}
    \end{small}
    \caption{}
    \label{fig:L3example2}
  \end{subfigure}
  \caption{\textbf{Q},\textbf{A} represents question and answer
    respectively. (a) shows the image followed by question at Level 1
    which is based on the low level features of image, without
    classification (b) has question at Level 2 which can be answered
    on applying a classifier (c),(d) shows questions which are based
    on scene text.}
  \label{fig:L1L2example}
\end{figure}

\item[Level 3 (L3):] L3 is different from the other levels and is
  based on recognising and processing scene text present in an
  image. There are many examples of scene text: sign boards, titles,
  clocks, calendars, diaries, etc. are a few such examples.
  \cref{fig:L3example1,fig:L3example2} show examples of questions that
  come under L3.

\item[Level 4 (L4):] From this level questions asked may require
  knowledge that may not be directly present as visual
  information. Questions which require knowledge of \emph{a single}
  object for answering are in Level 4. In other words, these are
  questions asked about the properties (listed on Page
  \pageref{objprops}) of a single object previously classified by a
  classifier. Some examples are shown in \cref{fig:examplel4}.

    An important subclass of questions in L4 are of the type ``which
    game/sport''. These are placed in L4 because identifying a
    sport/game is generally based on a distinguishing object such as a
    tennis racqet, hockey stick, golf club, \etc.

\begin{figure}[tb]
  
   \begin{center}
     \begin{tabular*}{\textwidth}{@{\extracolsep{\fill}}p{.5cm}p{3.6cm}p{3.6cm}p{3.6cm}}

    & \includegraphics[width=3.4cm,height=3.4cm]{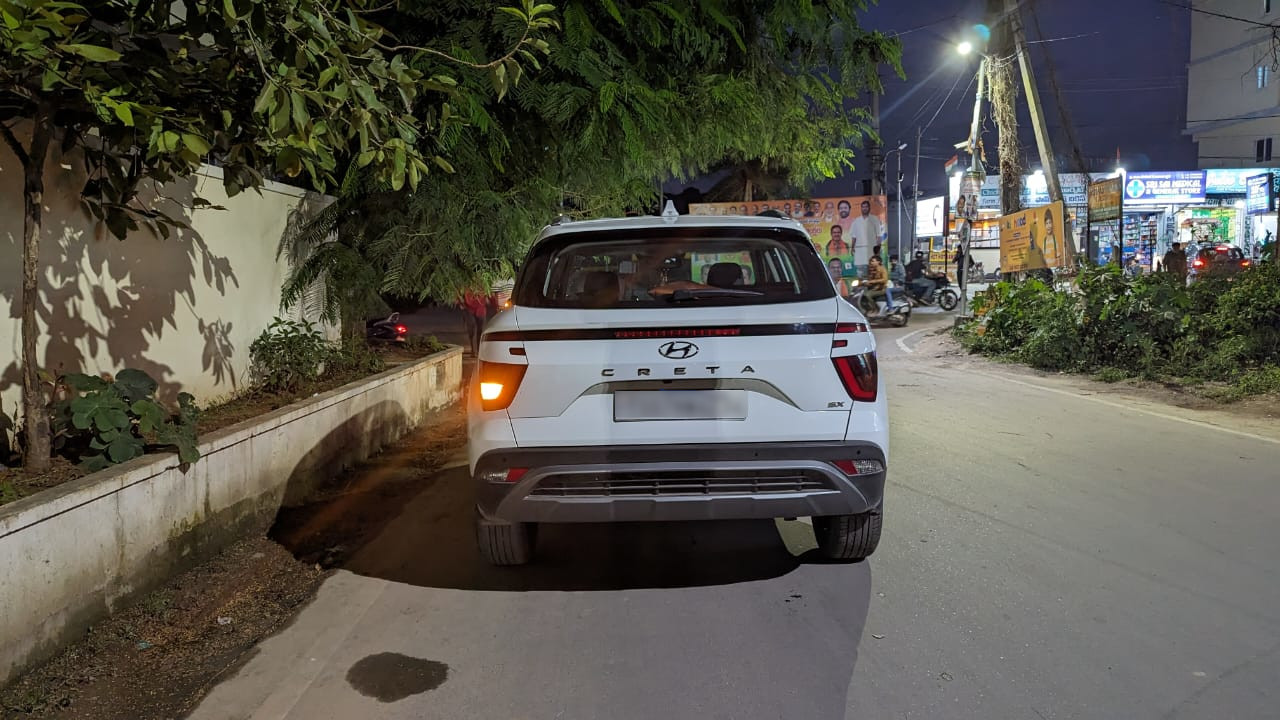} &
    \includegraphics[width=3.4cm,height=3.4cm]{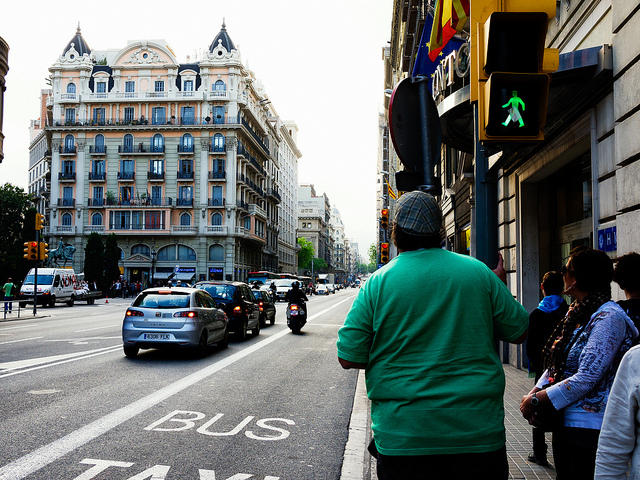}&
    \includegraphics[width=3.4cm,height=3.4cm]{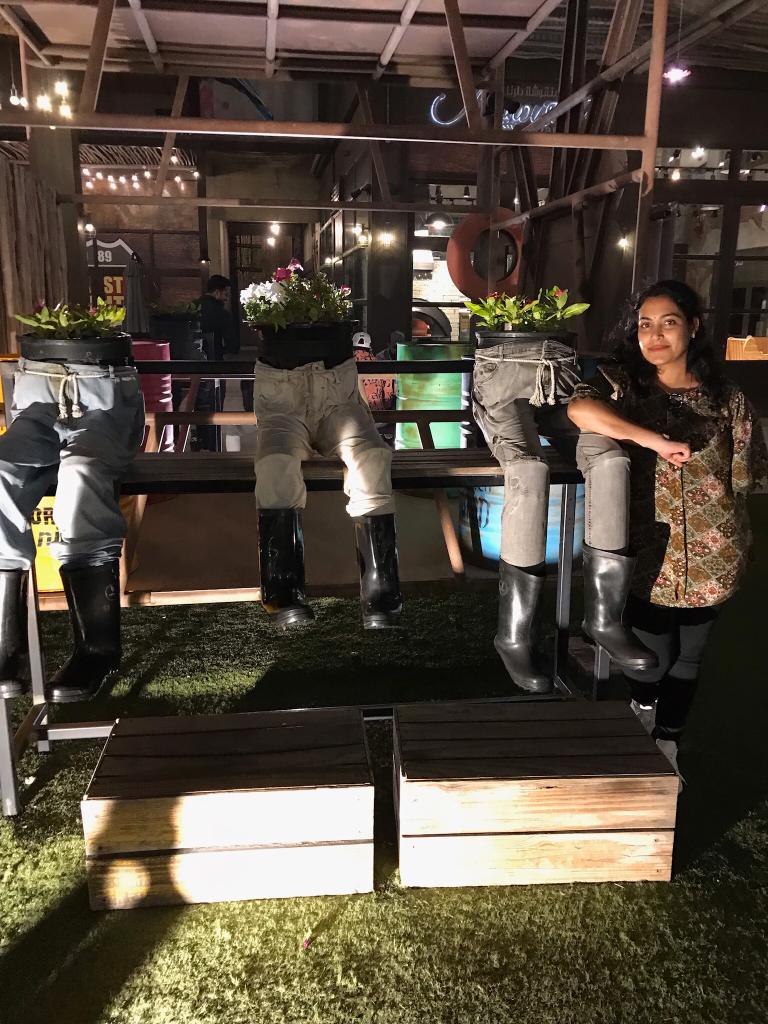}
     \\
    
     \textbf{Q} &  In which direction will the vehicle turn? & Can the
     pedestrians cross the road?& Is the person angry?\\
     \textbf{A}  &  left & yes & no\\
     \textbf{P} &  direction & traffic sign & emotions \\
     \textbf{O}   &  left indicator  & walk signal  & single person\\
    &
    \includegraphics[width=3.4cm,height=3.4cm]{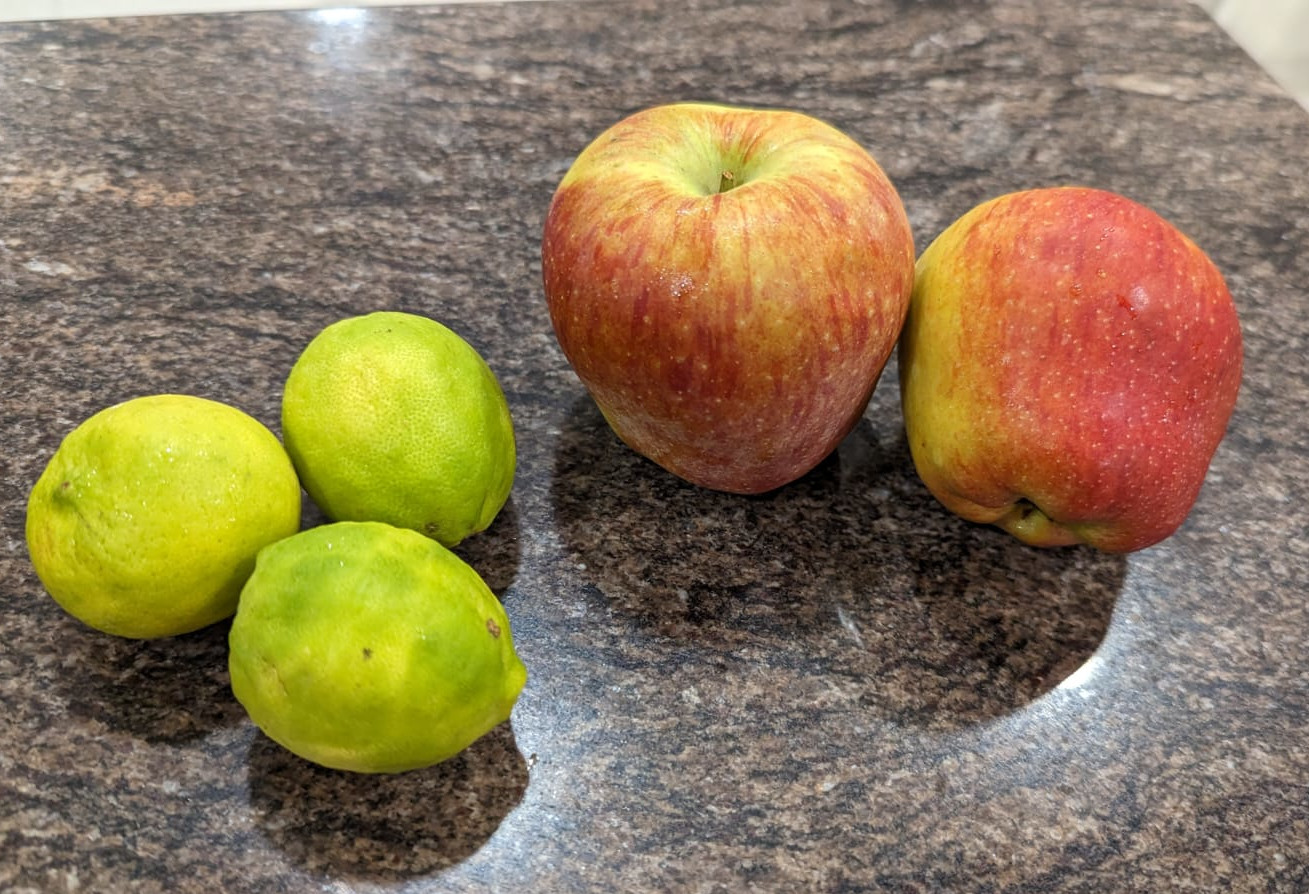} &
    \includegraphics[width=3.4cm,height=3.4cm]{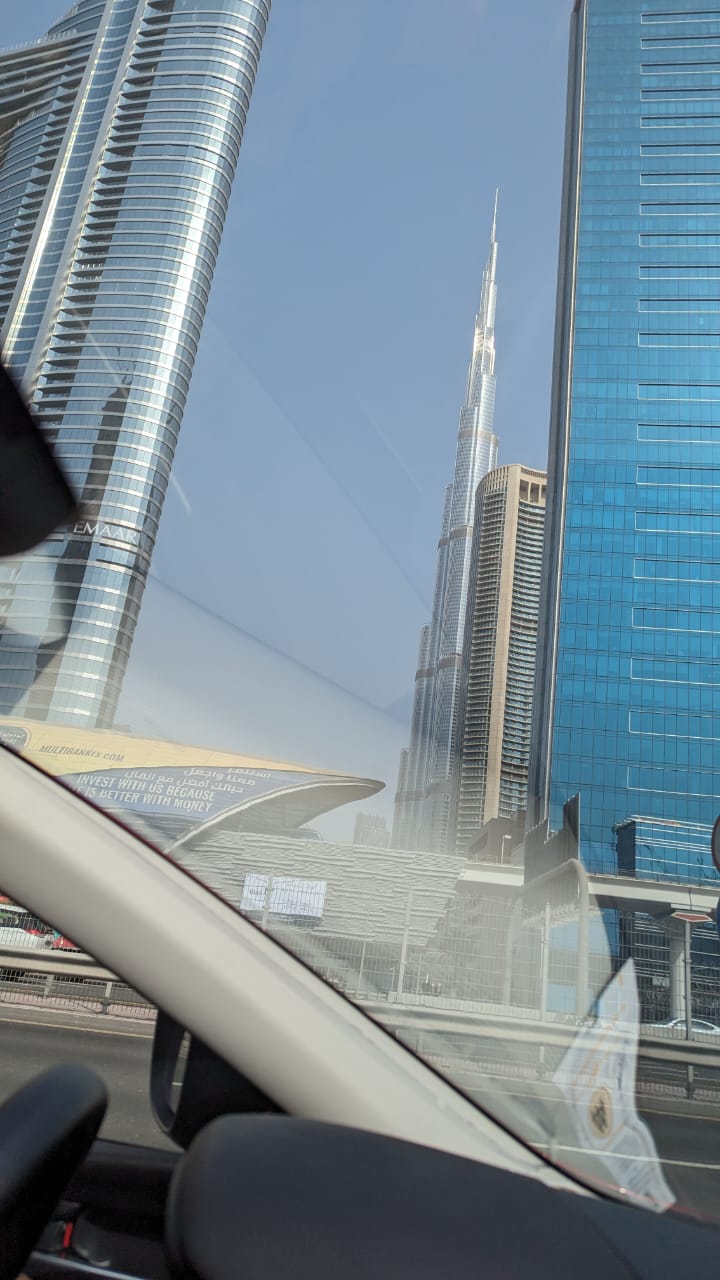}
     \\
     \textbf{Q} 
              & Which fruit tastes sour? & Which city is this?
    \\
     \textbf{Q} &  lime & Dubai\\
     \textbf{P}  & intrinsic property & distinguishing property\\
     \textbf{O}  &  of limes & Burj Khalifa
    \\
    \end{tabular*}
    \caption{Level 4 questions which require analysis of single object
      for answering, here \textbf{Q} is question, \textbf{A} is
      answer, \textbf{P} is property on which the question is based
      on, \textbf{O} is object from which we are getting answer. }
    \label{fig:examplel4}
    \end{center}
    
    \end{figure}

\item[Level 5 (L5):] Questions based on the features listed on
      Page \pageref{objprops} but requiring analysis of more than one
      object and their interactions. Examples are in 
  \cref{fig:examplel5}.

    \begin{figure}[tb]
    
   \begin{center}
    
    \begin{tabular*}{\textwidth}{@{\extracolsep{\fill}}p{.5cm}p{3.6cm}p{3.6cm}p{3.6cm}}

    &\includegraphics[width=3.4cm,height=3.4cm]{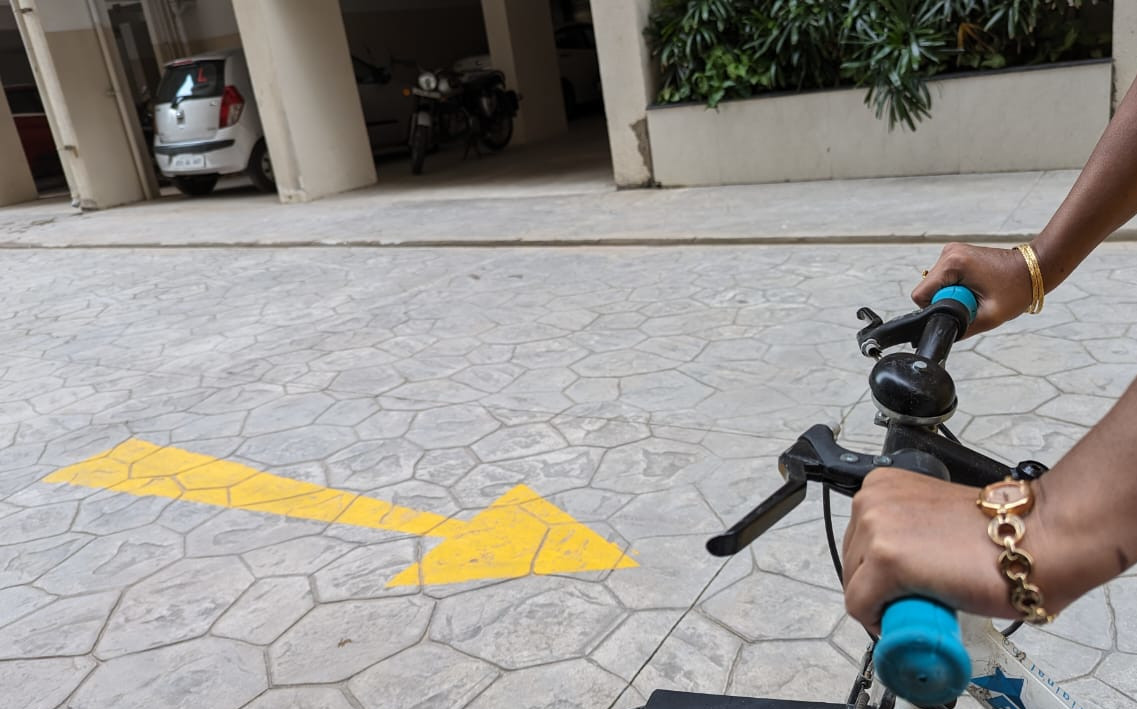}
    & \includegraphics[width=3.4cm,height=3.4cm]{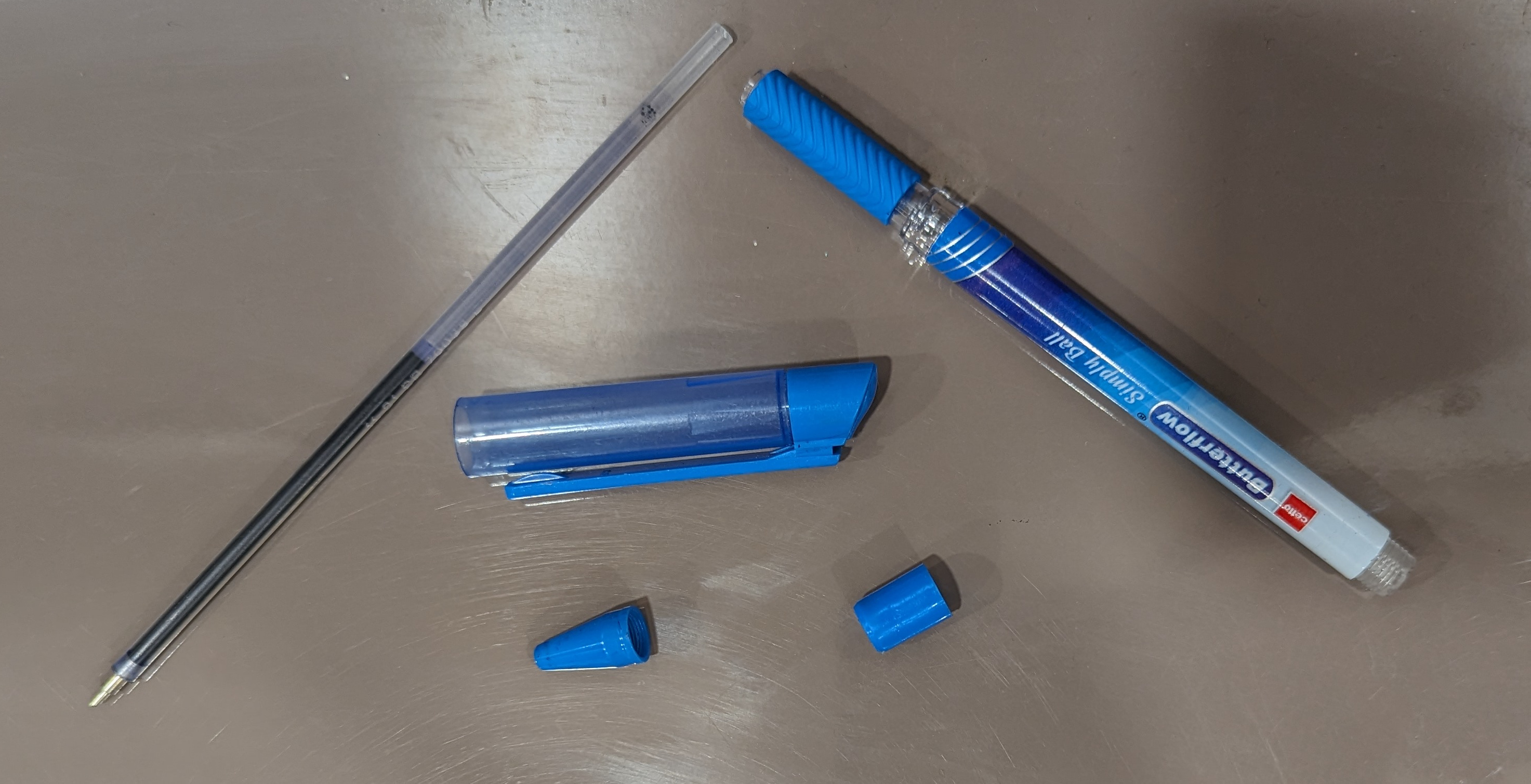}
    &  \includegraphics[width=3.4cm,height=3.4cm]{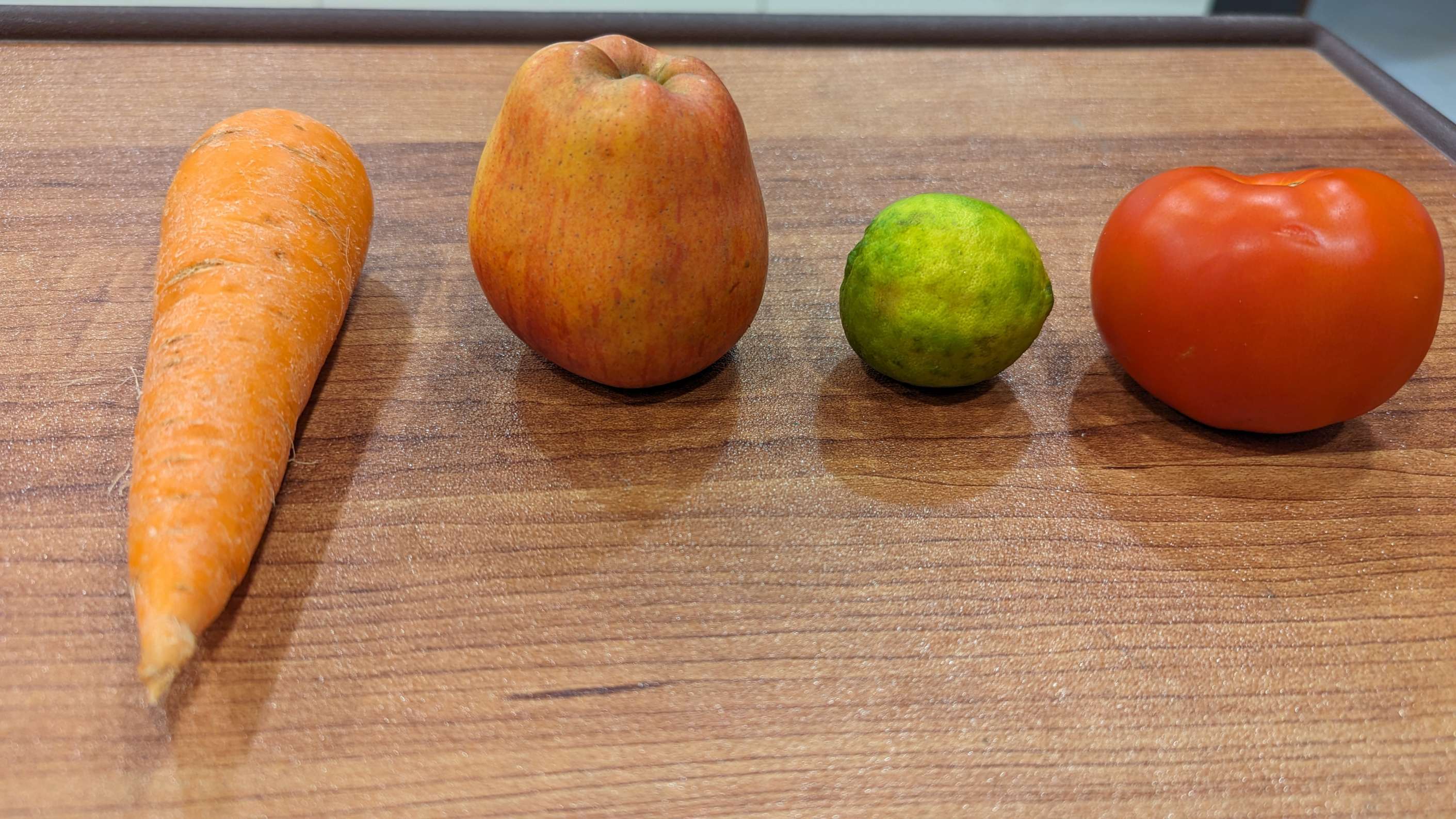}
     \\
    
    \textbf{Q} & Is the person riding on the correct side of the road? &
    From which object did the parts shown in the image come? & Which fruit
    shown in the photo is the sweetest?
    \\
    \textbf{A}  & no & pen & apple
    \\
    \textbf{O}   & arrow and cycle directions & different parts of
    pen & comparison

    \end{tabular*}
    \caption{Level 5 questions requiring analysis of multiple objects
      here \textbf{Q} is question, \textbf{A} is answer, \textbf{O}
      are objects from which we are getting answers.}
    \label{fig:examplel5}
    \end{center}
    
    \end{figure}

\item[Level 6 (L6):] Questions on extrapolated, hidden and occluded
  objects come under Level 6. For example, three players from a mixed
  doubles match are shown in the image and the question asks the
  gender of the fourth person as in \cref{fig:L6examplea}. A car
  dashboard is shown and the question is on the indicator signals at
  the back of the vehicle (which are not seen in the image)
  \cref{fig:L6exampleb}.

\begin{figure}[tb]
  \centering
  \begin{subfigure}{0.48\linewidth}
    \centering
    \includegraphics[width=0.8\textwidth]{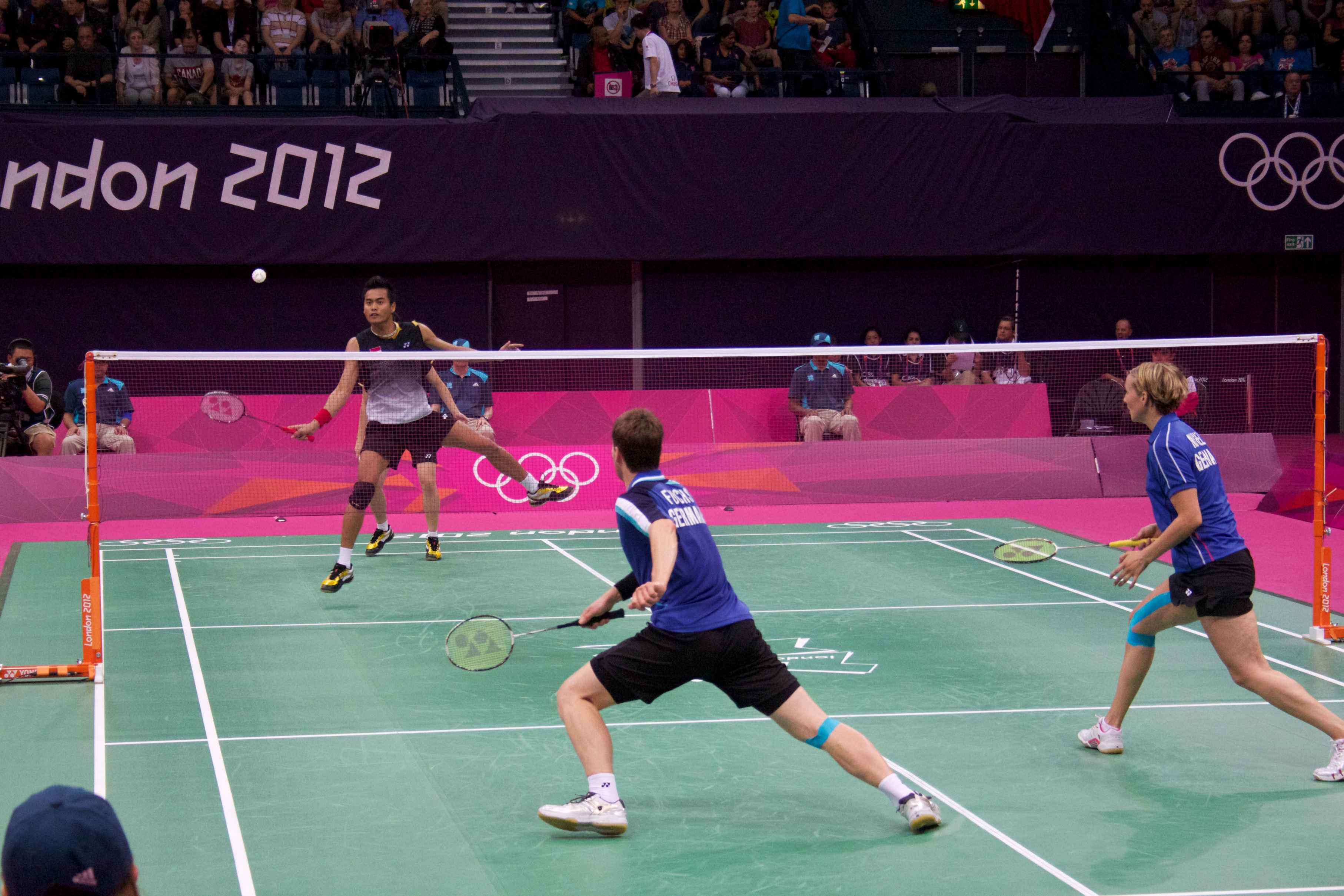}
    \vspace{\baselineskip} 
    \begin{small}
        \begin{tabular}{p{1.5cm}p{4cm}}
            
            \textbf{Question}: & What is the gender of the occluded fourth person on
     the far side of the net?\\
            \textbf{Answer}: & female    \\
                       
        \end{tabular}
    \end{small}
    \caption{}
    \label{fig:L6examplea}
  \end{subfigure}
  \hfill
  \begin{subfigure}{0.48\linewidth}
    \centering
    \includegraphics[width=0.9\textwidth]{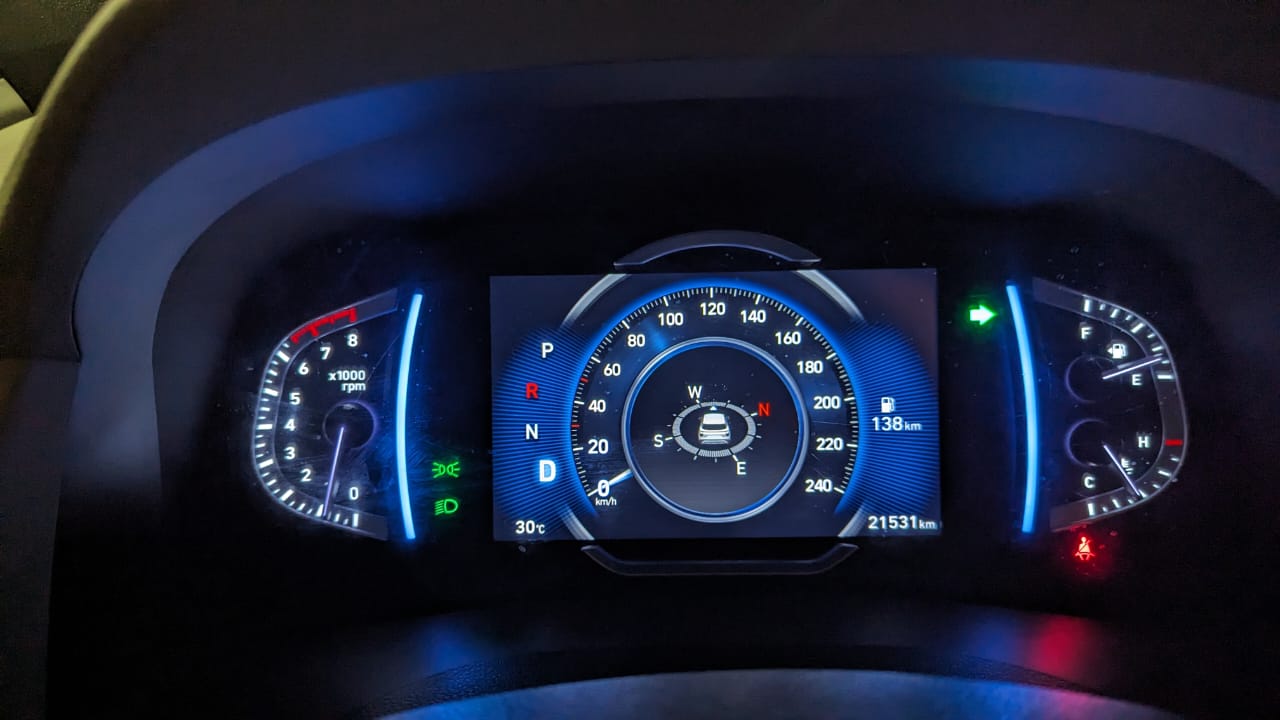}
    \vspace{\baselineskip} 
    \begin{small}
        \begin{tabular}{p{1.5cm}p{4cm}}
            \textbf{Question}: & Which turn indicator signal will be `ON' at the back of the vehicle?\\
            \textbf{Answer}: & right\\
            
        \end{tabular}
    \end{small}
    \caption{}
    \label{fig:L6exampleb}
  \end{subfigure}
  \caption{Questions at Level 6 on occlusion and extrapolation}
  \label{fig:L6example}
 \end{figure} 
 
\item[Level 7 (L7):] These questions require analysis of the entire
  image and sometimes beyond. The questions may be on intangible
  properties such as `happiness', `disasters', etc. Examples at this
  level can be ``What event is being celebrated?'', ``Why are the
  people hitting each other with tomatoes?'', etc as in \cref{fig:L7example}. 
\end{description} 
\begin{figure}[tb]
  \centering
  \begin{subfigure}{0.48\linewidth}
    \centering
    \includegraphics[width=0.5\textwidth]{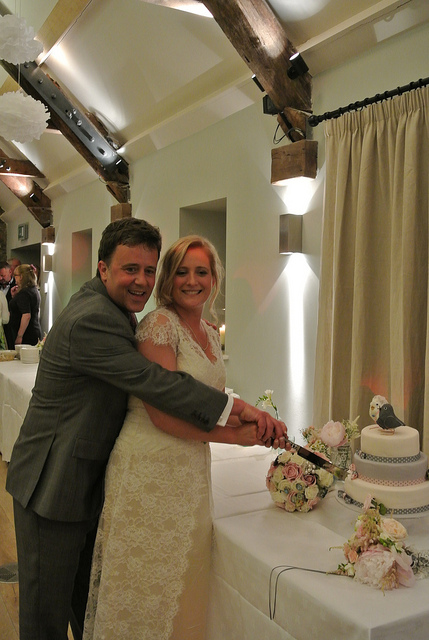}
    \vspace{\baselineskip} 
    \begin{small}
        \begin{tabular}{p{1.5cm}p{4cm}}
            
            \textbf{Question}: & Which event is celebrated?\\
            \textbf{Answer}: & wedding    \\
                       
        \end{tabular}
    \end{small}
    
  \end{subfigure}
  \hfill
  \begin{subfigure}{0.48\linewidth}
    \centering
    \includegraphics[width=\textwidth]{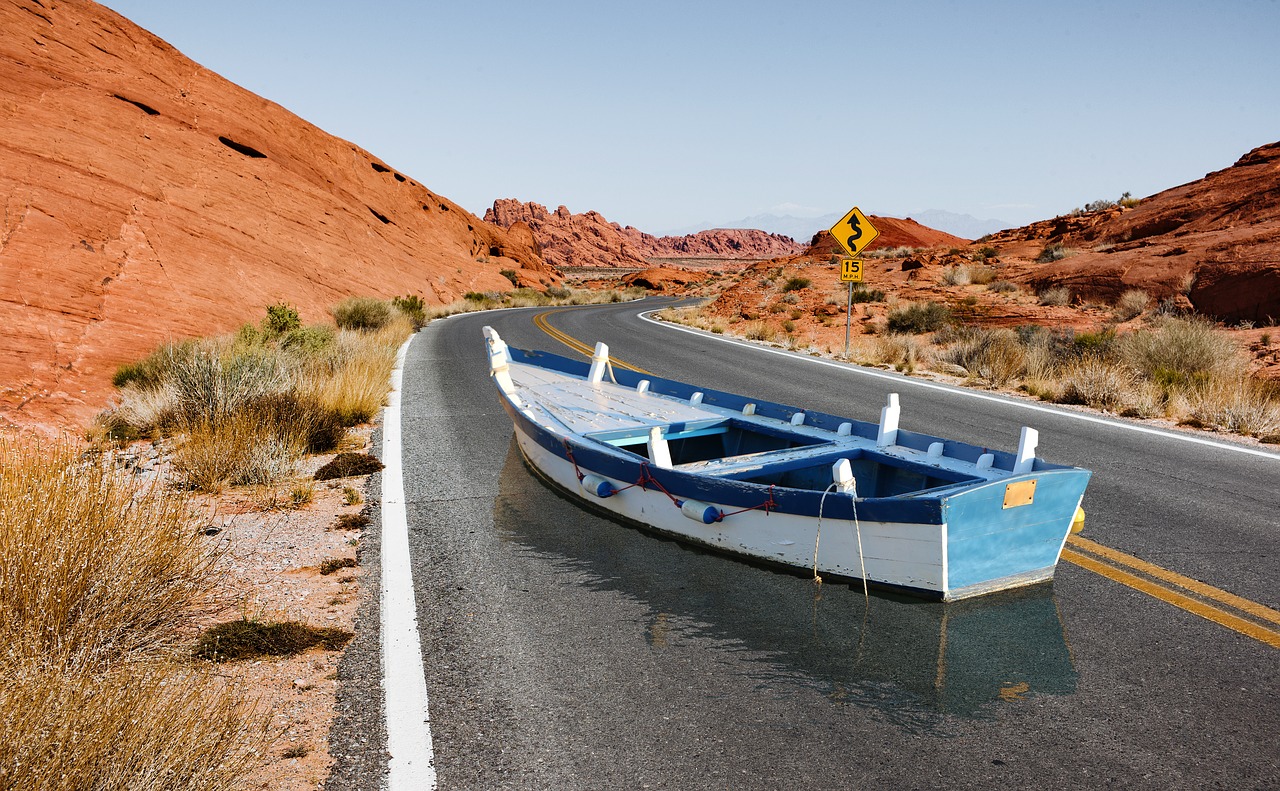}
    \vspace{\baselineskip} 
    \begin{small}
        \begin{tabular}{p{1.5cm}p{4cm}}
            \textbf{Question}: & Is the scene real?\\
            \textbf{Answer}: & no\\
            
        \end{tabular}
    \end{small}
    
  \end{subfigure}
  \caption{Questions at Level 7 on whole scene analysis and abstraction}
    \label{fig:L7example}
\end{figure}

The 7 levels of questions proposed in our dataset and their
definitions are summarised in  \cref{tab:summary}.
\begin{table}
\caption{Summarising the 7-level hierarchy}
  \label{tab:summary}
 
  \begin{center}
  \begin{small}
  \begin{tabular}{@{}lp{8.4cm}@{}}
    \toprule
   \textbf{Level}  & \multicolumn{1}{c}{\textbf{Criteria}} \\
    \midrule
     L1 & on low-level visual features (Marr's levels 1 and 2)\\
     L2 & L1 + classification of single and multiple objects\\
     L3 & scene text, clock time, calendar  \\
     L4 & L2 + Features listed in Page \pageref{objprops} on a single object \\
     L5 & L4 on multiple objects and their interactions\\
     L6 & L1, L2, L3, L4 + extrapolation to occlusions \\
     L7 &  analysis of complete scenes including intangible properties\\
    
    \bottomrule
  \end{tabular}
  \end{small}
\end{center}

\end{table}
\section{Benchmark dataset for testing VQA systems}
A dataset \textbf{VQA-Levels} with 210 images and 751 questions has
been created as a pilot project. 170 images were taken from the
Microsoft Common Objects in Context (MS COCO) dataset
\cite{lin2014microsoft}, 9 images were taken from the web and the
remaining 31 images are from the authors. 

A user interface was created to help in creating the dataset. In
question \emph{generation mode}, the interface randomly selected an
image from a database of about 10,000 images and presented it on the
screen. The user was given an option either to enter questions or skip
to another image. When entering the questions, the user also had to
select the level of the question and the reason for doing so. The user
also had to enter an answer. The question levels were also displayed
in the interface to assist the users. A group of around 50 people
consisting of graduate and Ph.D. students volunteered in creating the
dataset. They were trained on the user interface and the question
levels.  

These student volunteers were also provided with a \emph{question
  answering} mode. In this mode, images for which questions were
earlier given, are shown to the user along with the questions. The
user was given the space to answer the question. These images are not
the ones for which they framed the questions. Once the user answered
the question, the answer originally given was displayed on the
screen. The user was asked whether he/she agreed with the answer or
wishes to change the answer to his/her choice.

A total of about 1300 questions on 400 images were entered by the
volunteers. These questions and answers have been evaluated for
correctness, were filtered and a subset of them have been selected for
inclusion in the dataset. All the questions and annotations are human
monitored.

\subsection{Dataset Statistics}
In the VQA-Levels dataset consisting of 210 images and 751 questions
there are a minimum of 2 and maximum of 8 questions questions on each
image. Questions are close to what a human would ask when seeing the
image. They are also categorised into the levels L1 to L7 as defined
in the previous section. 

We found that there are four types of questions: yes/no (YN),
binary choice (BC), number (NO) and freestyle (FS). The focus is on
minimising ambiguity and multiple possible answers to a question. The
images chosen are also general and the questions asked may be answered
by anyone without being region or country-specific. There are very few
questions at L1 as humans do not seem to ask such questions. Most
questions are at L2 and L4 showing that humans tend to focus on one
object in the image.

The volunteers were also asked to minimise the YN and BC type
questions because our initial tests on the existing VQA systems showed
that they answer these questions by making a random choice. Their
accuracy on these types of questions was around 50\% showing that
these questions really do not indicate the true performance. 
Many such questions were found and filtered out from levels L5 and L7
where the volunteers gave a significant number of YN and BC
questions. Examples are questions such as, ``Is the person in
danger?'' or ``Is it a happy scene?'' In both the cases, there were
three ``Yes'' and three ``No'' answers to these types of questions
from the existing systems. Only which systems give a ``yes'' or a
``no'' answer changed between the questions.

Statistics of dataset for different levels and question types
is shown in \cref{tab:datstat}.
\begin{table}[tb]
\caption{VQA-Levels dataset statistics showing numbers of different
    types of questions for each level. Here BC, NO, FS, YN indicate binary choice, number, freestyle and yes/no questions respectively.}
  \label{tab:datstat}

  \begin{center}
  \begin{small}
  \begin{sc}
  \begin{tabular}{@{}lccccc@{}}
    \toprule
   \textbf{Level}  & \textbf{BC} & \textbf{NO} &\textbf{FS}  &
   \textbf{YN} & \textbf{Total} \\ 
    \midrule
     \textbf{1} & 0  &  0 &  10 &  0 & \textbf{10} \\
     \textbf{2} & 1  & 58 & 186 &  7 &\textbf{252} \\
     \textbf{3} & 3  & 16 &  73 &  1 & \textbf{93} \\
     \textbf{4} & 9  & 29 & 148 & 14 & \textbf{200} \\
     \textbf{5} & 5  & 19 &  74 & 16 & \textbf{114} \\
     \textbf{6} & 0 & 16 & 25 & 3 & \textbf{44} \\
     \textbf{7} & 2 & 4 & 27 &  5 & \textbf{38}\\
     \midrule
     \textbf{Total} & 20 & 142 & 543 & 46  & \textbf{751}\\
    \bottomrule
  \end{tabular}
  \end{sc}
  \end{small}
\end{center}

\end{table}
\section{Experiments and Results}
\label{sec:expandres}
We want to evaluate the performance of existing VQA-models at each
level rather than the overall performance. We have tested the
performance of the following VQA models on VQA-Levels dataset
\begin{itemize}
    \item   Pythia v0. 1\cite{jiang2018} which was the winning entry
      in the 2018 VQA challenge, it is a reimplementation of up-down model
        \cite{anderson2018,teney2018} with modularity
    \item   Vision-and-Language Transformer (ViLT)\cite{kim2021}: both
      textual inputs and visual inputs are processed free of
      convolution.
    \item   One For All (OFA)\cite{wang2022} which integrates a
      variety of unimodal and cross-modal tasks into a simple
      sequence-to-sequence learning framework using single instruction
      based task representation
    \item   Bootstrapping Language Image Pre-training for unified
      vision-language understanding and generation
      (BILP)\cite{li2022}. It uses a vision language pretraining
      framework that adapts easily to tasks involving both generation
      and vision language understanding
    \item   Generative Image-to-text Transformer
      (GIT)\cite{wang2022git} which uses a simplified architecture of
      one image encoder and one text decoder. 
\end{itemize}

The level-wise performance of these models is shown in Table
\ref{tab:resL1L7}. Their overall performance is in the 50\% range
while the level-wise performance varies between 10\% and 90\% showing
that the question complexity plays an important role. When L3 is
discarded from the analysis, the performance varies from 45\% to
90\%.
\begin{table} [tb]
\caption{Performance of VQA models on L1 to L7 questions and overall accuracy.}
  \label{tab:resL1L7}

   \begin{center}
   \begin{small}
  \begin{tabular}{@{}lccccccc@{}}
    \toprule
  \textbf{Level}   & \textbf{Pythia} & 
            \textbf{ViLT}  &
            \textbf{OFA} &
            \textbf{GIT\_base} & 
             \textbf{GIT\_large} &
              \textbf{BLIP\_base} &
               \textbf{BLIP\_large} \\
    \midrule
    \textbf{1} & 40.0 & 80.0& 70.0 & 50.0& 60.0&90.0 & 90.0\\
    \textbf{2} & 52.59 & 58.17 & 62.55 & 47.01&55.38 & 68.92& 68.53\\
    \textbf{3} & 16.13 & 12.9 & 17.2&17.2&29.03 &20.43 &18.28\\
    \textbf{4} & 44.5 & 48.5& 55.5 & 44.5& 54.5& 61.0&61.0\\
    \textbf{5} & 49.12 &50.88 & 49.12 &48.25 & 56.14 & 57.89&57.89\\
    \textbf{6} & 45.45 & 40.91& 43.18 & 31.82& 40.91& 40.91&40.91\\
    \textbf{7} & 26.32 & 39.47 & 39.47 &31.58 &47.37 & 47.37&47.37\\
    \midrule
    \textbf{Overall} & 43.41  & 47.14& 50.87 & 41.15&50.73 &56.72 &56.32\\
    \bottomrule
  \end{tabular}
  \end{small}
\end{center}

\end{table}

\section{Discussion}
\label{discussion}

Results in Table \ref{tab:resL1L7} show some significant points of
interest. First, the general trends indicate that the models performed
best on L1 and L2 questions as expected. These questions are directly
related to the visual content in the images. The least performance is
on Level L3, then L6 and L7 questions. This is a vindication of the
levels proposed in our dataset that the questions are categorised
according to their difficulty. L3 is, as expected, an anomaly as it
requires special training for detecting and recognising scene text.

\begin{figure}[tb]
  \centering
  \begin{subfigure}{0.45\linewidth}
    \centering
    \includegraphics[width=0.65\textwidth,height=2.8cm]{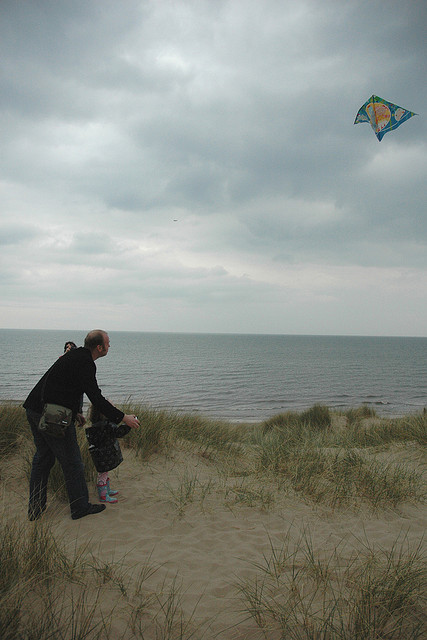}
    \vspace{\baselineskip} 
    \begin{small}
        \begin{tabular}{p{.5cm}p{5.2cm}}
            
            \textbf{Q}: & What connects the kite to the child?
\\
            \textbf{A}: &  thread, string
\\
            \textbf{L}: &  6
\\       
           
        \end{tabular}
    \end{small}
    \caption{}
    \label{fig:discussionexamplea}
  \end{subfigure}
  \hfill
  \begin{subfigure}{0.45\linewidth}
    \centering
    \includegraphics[width=0.65\textwidth,height=2.8cm]{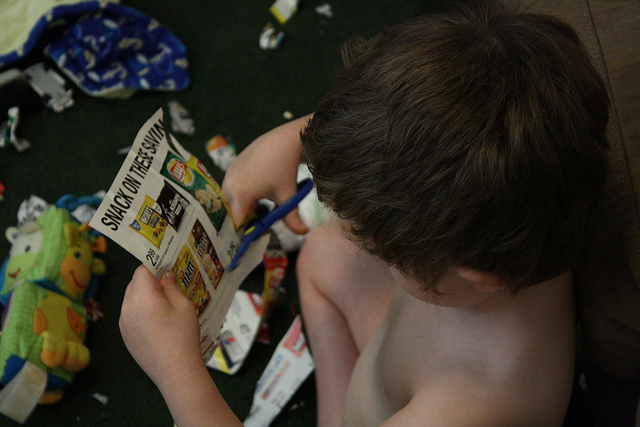}
    \vspace{\baselineskip} 
    \begin{small}
        \begin{tabular}{p{.5cm}p{5.2cm}}
           
            \textbf{Q}: & Why is the man cutting the paper?
\\
            \textbf{A}: & for coupons
\\
	\textbf{L}: &  7
\\

        \end{tabular}
    \end{small}
    \caption{}    
    \label{fig:discussionexampleb}
  \end{subfigure}
  
  \caption{\textbf{Q},\textbf{A}, \textbf{L} represents question, answer and level respectively. (a) shows question on invisible object (b) require analysis of the whole scene.}
  \label{fig:discussionexample}
\end{figure}

Second, the VQA systems today are \emph{using the information of
  language models than the vision based semantics}.
\cref{fig:discussionexamplea} shows an image where the child is flying
a kite and thread is not visible in the image. The question asked is,
``What connects the kite to the child?". Almost all the systems tested
answered correctly as ``string.'' But, a very similar question, ``What
is the child holding in her left hand?'' on another image of a child
flying the kite resulted in a wrong answer from all the systems.  When
asked the question ``Why is the man cutting the paper?"  on
\cref{fig:discussionexampleb} whole scene analysis is required to give
the answer, ``for coupons''. No system tested gave the correct
answer. They all answered either ``cut'' or ``to open'' the
paper. Questions at these levels proved difficult unless there is a
\emph{word closely related to the answer} in the question. However,
performance at Levels L6 and L7 requires greater experimentation as
the numbers of questions at these levels is low.

Third, the dominance of language models is further illustrated on
certain questions at Levels L4 and L5. When asked questions about directions
like ``In which direction is the wind blowing?  " on the image in
\cref{fig:discussionexamplea} models give answers such as
East, West, sometimes even North, which is not the natural way for humans
to answer such questions. Most of the volunteers correctly gave the
answers as ``from left to right'' or ``from the left''. In fact none
of them gave the answers as ``East, West,'' \etc.

Fourth, absurd answers are given: again as expected because these
systems have no causal or semantic understanding. For example, on an
image of a partially eclipsed Moon, the question, ``What is in the
image?'' resulted in the incorrect answer of ``plate.'' The question,
``Does it show an eclipse?'' led to two ``no'' and four ``yes''
answers. The VQA systems, of course, lack the knowledge that a plate
cannot be eclipsed.

Finally, there is a significant need to work on scene text and use it for
answering questions. The systems tested could answer questions
correctly on prominent text in the picture  but had
difficulty in clock times, diary-based,  or calendar-based questions. 

Our goal is to have at least 250 questions in each level. L2 and L4
questions have reached close to that count. More number of questions in
L6 and L7 have to be added. Questions in these levels have to be
manually added and it is also difficult to find images where humans naturally
ask such abstract questions. Question generation at L1 can be
automated as they are on the basic features of image. L2 questions can
be automatically generated based on objects identified and their
presence, count and color. We are in the process of generating
questions at Levels L4 and L5 automatically by using the
classification labels as keywords in Internet search. Automating
generation of L6 and L7 questions appears difficult with the current
systems.

From the above, we see that current VQA systems, built by combining
visual recognition and language models perform well on questions
involving visual content and image description. Their performance
drops when questions that require external knowledge, extrapolation to
unseen parts of the scenes, abstraction, \etc. This is especially the
case when the questions do not contain any words that are related to
the answers.

\section{Conclusion}
\label{conclusion}
In this paper, we created a new VQA-Levels dataset that contains
questions systematically arranged into 7 levels based on their
complexity and the level of analysis needed for answering. These
levels are inspired by Content Base Image Retrieval and David Marr's
theory of vision. The pilot dataset, although small, revealed
interesting differences in level-wise performance when tested on
state-of-the-art VQA systems. Analysis of these results shows that
today's VQA systems rely more on language models than on visual
semantics. The results also show where immediate impact may be
achieved in developing new VQA systems: on scene text analysis. The
results also show that VQA systems need to be built with stronger
vision engines. The level wise performance results also show that our organization of questions into 7 levels is valid reflection of complexity levels.

%
%
\bibliographystyle{splncs04}
\bibliography{main}
\end{document}